%% file: main.tex
\documentclass{article}

\usepackage{microtype}
\usepackage{graphicx}
\usepackage{subcaption}
\usepackage{booktabs} 
\usepackage[most]{tcolorbox}
\usepackage{minted} 
\usepackage{tcolorbox}
\tcbuselibrary{breakable}
\usepackage{multirow}
\usepackage{comment}

\usepackage{hyperref}
\usepackage[table]{xcolor}

\usepackage[preprint]{icml2026}

\usepackage{amsmath}
\usepackage{amssymb}
\usepackage{mathtools}
\usepackage{amsthm}

\usepackage[capitalize,noabbrev]{cleveref}

\theoremstyle{plain}

\theoremstyle{definition}

\theoremstyle{remark}

\usepackage[textsize=tiny]{todonotes}

\icmltitlerunning{CoRPO: Adding a Correctness Bias to GRPO Improves Generalization}

\begin{document}
\twocolumn[
  \icmltitle{CoRPO: Adding a Correctness Bias to GRPO \\ Improves Generalization}




\begin{icmlauthorlist}
  \icmlauthor{Anisha Garg}{cbs}
  \icmlauthor{Claire Zhang}{cbs}
  \icmlauthor{Nishit Neema}{cbs}
  \icmlauthor{David Bick}{cbs}
  \icmlauthor{Ganesh Venkatesh}{comp}
  \icmlauthor{Joel Hestness}{cbs}
\end{icmlauthorlist}


\icmlaffiliation{cbs}{Cerebras Systems}
\icmlaffiliation{comp}{Work done while at Cerebras Systems}
\icmlcorrespondingauthor{Anisha Garg}{anisha.garg@cerebras.net}


  \icmlkeywords{Machine Learning, ICML}

  \vskip 0.3in
]

\printAffiliationsAndNotice{}  

\newcommand{\fixme}[1]{\textcolor{red}{#1}}
\newcommand{\ignore}[1]{}
\newcommand{\corpo}{\textsc{CoRPO}}
\newcommand{\corpof}{Correctness Relative Policy Optimization (\textsc{CoRPO})}

\input{0.abstract}

\input{1.introduction_v2}
\input{2.background}
\input{3.grpo-challenge}

\input{4.corpo}
\input{5.results-v2}

\input{6.conclusion}

\bibliographystyle{unsrtnat}
\bibliography{refs}

\input{7.appendix}

\end{document}

%% file: 0.abstract.tex
\begin{abstract}

Group-Relative Policy Optimization (GRPO) has emerged as the standard for training reasoning capabilities in large language models through reinforcement learning. By estimating advantages using group-mean rewards rather than a learned critic, GRPO has enabled efficient scaling of reinforcement learning from verifiable rewards (RLVR). However, we identify a fundamental limitation: GRPO's mean baseline can assign positive advantages to incorrect solutions simply because they outperform a poorly-performing group average. It leads to overestimation of advantages and reinforcement of incorrect behaviors.

To address this, we propose Correctness-Relative Policy Optimization (CoRPO), a simple modification to the GRPO objective that clips the minimum baseline to a fixed correctness threshold. We show that baseline clipping introduces a protective bias to advantage estimation that mitigates overfitting while preserving effective exploration.  Empirically, CoRPO-trained models improve cross-domain reasoning, generalizing more consistently to out-of-domain (OOD) tasks. When trained on coding tasks, CoRPO outperforms GRPO on math, and vice-versa, indicating that CoRPO learns robust, transferable reasoning patterns rather than task-specific solutions.
\end{abstract}

%% file: 1.introduction_v2.tex
\section{Introduction}
\label{sec:intro}

RLVR has emerged as a powerful paradigm for improving large language models on tasks with automatic correctness signals, such as mathematics and code generation~\cite{openai_deep_research_2025,fu2025deepthinkconfidence}. Group-Relative Policy Optimization (GRPO)~\cite{shao2024deepseekmathpushinglimitsmathematical} has become the de facto standard for RLVR, replacing learned value functions with the average 
reward of sampled trajectories as a baseline. This design significantly improves 
computational efficiency over PPO~\cite{schulman2017proximalpolicyoptimizationalgorithms}.

Despite its practical success, we identify two fundamental limitations in GRPO's 
baseline construction. First, GRPO estimates the policy's expected reward using 
the empirical mean of a small sample (typically 4–16 rollouts)~\cite{khatri2025art}. When this 
estimate falls below the true expectation, which occurs with non-trivial 
probability, average or even suboptimal trajectories receive inflated positive 
advantages, leading to overly aggressive updates. Second, when rewards are 
ordinal~\cite{sun2025rl, zhang2026chaining} and non-calibrated, common in LLM-as-a-judge scenario~\cite{liu2504inference}, GRPO's group-mean baseline can assign 
positive advantage to \emph{incorrect} trajectories simply because they 
outperform other failures in the group. This fundamentally inverts the desired 
learning signal, directly reinforcing failed behaviors.

\begin{figure*}[h!]
    \centering
     \includegraphics[width=\textwidth]{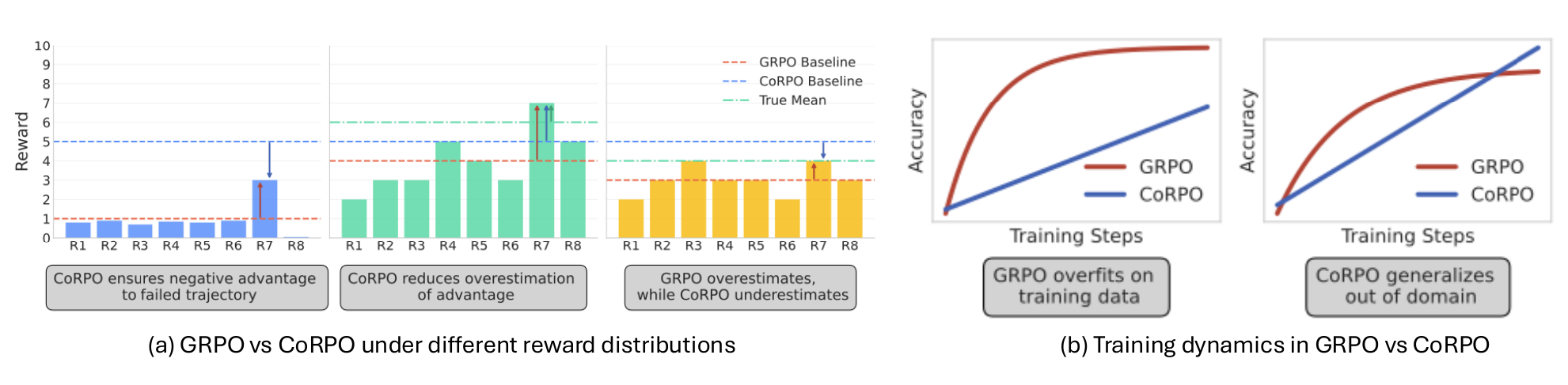} 
    \caption{
\textbf{Left:} Three scenarios comparing true mean reward to GRPO and CoRPO baselines. Arrows indicate advantages relative to the respective baselines and follow the same color coding as the baselines. (i) CoRPO ensures a trajectory receives negative advantage if it falls below the minimum correctness threshold. (ii) CoRPO reduces advantage overestimation when the true mean exceeds the correctness threshold but the sample mean is lower. (iii) CoRPO eliminates overestimation entirely when the correctness threshold exceeds the true mean, which in turn is above the sample mean. \textbf{Right:} By reducing advantage overestimation, CoRPO improves out-of-domain performance, even if GRPO achieves higher performance on the training distribution. Figure shows simulated data.
}
    \label{fig:grpo_corpo_baselines}
    \vspace{-10pt}
\end{figure*}

We propose \textbf{Correctness-Relative Policy Optimization (CoRPO)}, which 
addresses both issues through a single mechanism: clipping the group-mean 
baseline at a minimum correctness threshold (Figure ~\ref{fig:grpo_corpo_baselines}). This ensures that incorrect 
trajectories never receive positive advantage, regardless of variance in group composition due to stochastic generations. We show theoretically that this baseline clipping 
mitigates advantage overestimation and prevents premature exploitation. 
Empirically, on coding and mathematical reasoning tasks with ordinal rewards, 
CoRPO consistently outperforms GRPO on OOD evaluations, demonstrating 
superior cross-domain generalization. Analysis of training dynamics reveals 
that CoRPO overcomes GRPO's distribution sharpening by learning primarily 
through negative reinforcement of incorrect behaviors.

Our contributions are: (1) identifying and analyzing two failure modes of GRPO's 
baseline, advantage overestimation from variance in group composition and sign inversion 
under ordinal rewards; (2) proposing CoRPO, a simple baseline modification that 
addresses both issues while preserving GRPO's efficiency; and (3) demonstrating 
that CoRPO's correctness guarantee improves generalization and mitigates 
exploitative training dynamics.

%% file: 2.background.tex
\section{Background and Related Work}


\subsection{Policy Optimization for RLVR}

Proximal Policy Optimization (PPO)~\cite{schulman2017proximalpolicyoptimizationalgorithms} has long been the dominant algorithm for reinforcement learning with neural policies. PPO relies on a learned value function to estimate advantages, which stabilizes updates but introduces substantial computational and memory overhead. In the context of large language models~\cite{brown2020language, comanici2025gemini}, training an accurate value function, often comparable in size to the policy itself, is particularly challenging. Group-Relative Policy Optimization (GRPO)~\cite{shao2024deepseekmathpushinglimitsmathematical} addresses this challenge by eliminating the value function entirely. Instead, GRPO estimates advantage by subtracting the mean reward of a group of sampled trajectories for the same prompt. This design drastically reduces training cost and simplifies implementation, contributing to its rapid adoption in RLVR~\cite{guo2025deepseek} settings such as mathematical reasoning and code generation~\cite{jaech2024openai, lambert2024tulu, team2025kimi}.

However, measuring performance only relative to sampled peers rather than true expected returns introduces subtle but important biases. While group-relative formulation aligns well with preference-based reward models~\cite{zhu2025reinforcement}, it causes systematic errors when rewards are intended to reflect objective correctness.


\subsection{Use of baselines in Policy Gradient Algorithms}

Baselines in policy gradient methods are traditionally motivated as variance reduction tools~\cite{hankyang_policygradients, sutton2018reinforcement}. However, recent theoretical work challenges this interpretation in the context of large-scale language model optimization. In particular, it has been shown that the variance of natural policy gradient estimators can remain unbounded with or without a baseline, and that variance reduction is neither necessary nor sufficient for convergence~\cite{mei2022role}. Instead, baselines primarily function to control the \emph{aggressiveness} of policy updates. 

\subsection{Ordinal Rewards in RL for LLMs}

Although GRPO was originally proposed for binary rewards, recent work increasingly applies it to richer, ordinal rewards signals such as graded correctness or quality ratings~\cite{sun2025rl, zhang2026chaining}, to ensure that rewards can convey more information than just binary signals.

When combined with a group-mean baseline, ordinal rewards shift learning from absolute correctness to relative ranking, allowing incorrect trajectories to receive positive advantage if they outperform peers, and potentially making GRPO-based training suboptimal.

\subsection{Exploration--Exploitation Tradeoff and Distribution Sharpening in GRPO}

GRPO is known to exhibit exploitative tendencies during learning~\cite{hao2025rethinking, deng2025token}. Recent work~\cite{he2025rewardingunlikelyliftinggrpo} shows that GRPO exhibits a rank bias: trajectories that are already likely under the current policy are preferentially reinforced, while rare trajectories are neglected. This leads to \emph{distribution sharpening}, where probability mass concentrates on a narrow set of solutions. Complementary work~\cite{zhu2025surprisingeffectivenessnegativereinforcement} shows that penalizing incorrect trajectories alone (Negative Sample Reinforcement, NSR) can be highly effective, improving performance across the entire Pass@k spectrum, while reinforcing only correct trajectories degrades diversity. These results suggest that suppressing incorrect behaviors is critical for sustained exploration.

GRPO’s positive reinforcement of failed trajectories when using ordinal rewards conflicts with this principle, contributing to premature exploitation and distribution collapse. 





%% file: 3.grpo-challenge.tex
\section{Challenge with the GRPO baseline}
As discussed in Section~\ref{sec:intro}, GRPO's group-relative formulation 
measures performance against sampled peers rather than absolute correctness. 
We now formalize this design choice and identify its failure modes in RLVR 
settings.
\label{sec:challenge}

\subsection{Baseline and Advantage Calculation}

In policy gradient methods, advantages measure how much better an action performs 
compared to the expected return. GRPO estimates this using a group-mean baseline 
rather than a learned value function. Given $G$ rollouts $\{y_1, \ldots, y_G\}$ 
sampled from policy $\pi_\theta$ for the same prompt, GRPO computes:
\begin{equation}
b_{\text{mean}} = \frac{1}{G} \sum_{i=1}^{G} R(y_i)
\label{eq:grpo_baseline}
\end{equation}

The advantage for rollout $y_i$ is then computed as:
\begin{equation}
A(y_i) = R(y_i) - b_{\text{mean}}
\label{eq:grpo_advantage}
\end{equation}

\noindent We omit the normalization term for clarity, as our analysis holds 
under various normalization schemes~\cite{shao2024deepseekmathpushinglimitsmathematical, 
yu2025dapoopensourcellmreinforcement, liu2025understandingr1zeroliketrainingcritical}.

In GRPO, $b_{\text{mean}}$ can be seen as an estimator of the true expected reward 
$\mu^* = \mathbb{E}_{y \sim \pi_\theta}[R(y)]$ , while using finite number of samples. The advantage 
$A(y_i)$ thus measures whether trajectory $y_i$ performs better or worse than 
the average sampled trajectory. While this group-relative design eliminates the cost of training a value function, 
it introduces two fundamental issues in RLVR settings, which we analyze in the 
following subsections.

\subsection{Policy Gradient Objective}

The GRPO objective maximizes the expected advantage of sampled trajectories. For a rollout $y_i = (t_{i,1}, \ldots, t_{i,T_i})$ of 
length $T_i$, the objective is:

\begin{equation} 
J(\theta) = 
\mathbb{E}_{y_{i} \sim \pi_\theta,\, i \in [1,G]}
\left[
A(y_i)
\right]
\end{equation}


This yields the following policy gradient under language modeling:
\begin{equation}
   \nabla_\theta J(\theta)
=
\mathbb{E}_{y_{i} \sim \pi_\theta,\, i \in [1,G]}
\left[
A(y_i)
\sum_{j=1}^{T_i}
\nabla_\theta \log \pi_\theta(t_{i,j} \mid t_{i,<j})
\right] 
\end{equation}

The sign and magnitude of $A(y_i)$ directly control the update direction: 
positive advantages increase the probability of generating trajectory $y_i$, 
while negative advantages decrease it. Crucially, since $A(y_i) = R(y_i) - b_{\text{mean}}$ 
from Eq.~\ref{eq:grpo_advantage}, the learning signal depends on whether $y_i$ 
outperforms the group average rather than whether it is objectively correct. Preference signal is also weak because the group average is not necessarily equal to actual expected reward, $\mathbb{E}_{y \sim \pi_\theta}[R(y)]$. 
This relative formulation creates systematic failure modes in RLVR settings, 
which we analyze next.

\subsection{Challenge 1: Advantage Overestimation from Sampling Variance}
\label{sec:challenge1}

GRPO's baseline $b_{\text{mean}}$ estimates the true expected reward 
$\mu^*$ using only $G$ samples (typically 
4--16). This introduces estimation error $\epsilon = b_{\text{mean}} - \mu^*$. 
While $\mathbb{E}[\epsilon] = 0$ (the sample mean is unbiased), individual 
samples exhibit variance: $b_{\text{mean}}$ can fall above or below $\mu^*$.

When $b_{\text{mean}} < \mu^*$, advantages become inflated:
\begin{equation}
A(y_i) = R(y_i) - b_{\text{mean}} = \underbrace{(R(y_i) - \mu^*)}_{\text{true advantage}} + \underbrace{(\mu^* - b_{\text{mean}})}_{\text{positive bias}}
\label{eq:advantage_overestimation}
\end{equation}

The bias term $(\mu^* - b_{\text{mean}}) > 0$ inflates all 
advantages in the group. This occurs with probability 
$P(b_{\text{mean}} < \mu^*) = 1/2$, assuming $b_{\text{mean}}$ is gaussian with expected reward $\mu^*$. Approximately half of all 
sampled groups experience advantage overestimation.

This has two consequences: (1) overly aggressive updates to increase token probabilities even for trajectories whose rewards equal the expected reward, and 
(2) suboptimal trajectories can receive positive advantages when $b_{\text{mean}} < R(y_i) < \mu^*$, particularly when rewards are graded rather than binary (i.e., $R(y_i) \notin \{0,1\}$ .

\subsection{Challenge 2: Positive Advantage for Failed Trajectories}
\label{sec:challenge2}

We now examine GRPO’s group-mean baseline when the rewards are ordinal. Let $y_f$ denote a \textbf{failed trajectory}, i.e., one whose objective reward indicates incorrectness. This can occur in RLVR settings where partial-credit or proxy rewards can be uncalibrated, even when a coarse notion of correctness (e.g., pass/fail) exists for the task. A common example is using an LLM-as-a-judge to score responses for objectively verifiable tasks (see example in Appendix ~\ref{sec:failedtrajectory}). For simplicity, we assume
\begin{equation}
R(y_f) < 0,
\end{equation}
though the argument does not require rewards to be strictly negative. In policy gradient methods, a failed trajectory should never receive a positive learning signal:
\begin{equation}
A(y_f) \le 0.
\end{equation}

Under GRPO, however, the sign of the advantage depends on whether the trajectory exceeds the group baseline:
\begin{equation}
A(y_f) > 0 \iff R(y_f) > b_{mean}.
\end{equation}

Thus, a failed trajectory receives positive advantage whenever it is ``less bad’’ than the group average:
\begin{equation}
b_{mean} < R(y_f) < 0.
\end{equation}

This scenario is common early in training or on difficult tasks, where most rollouts fail and the group baseline is strongly negative. In these cases, trajectories that are objectively incorrect can be reinforced simply because they outperform a poorly performing group.

%% file: 4.corpo.tex
\section{\corpof}
\label{sec:corpom}

\subsection{Desiderata for an Ideal Baseline in RLVR}
\label{sec:ideal_baseline}

The challenges identified in Section~\ref{sec:challenge} motivate a 
reconsideration of baseline design for RLVR. We argue that an ideal baseline 
for policy optimization in settings with verifiable rewards should satisfy three 
key properties:
\begin{enumerate}
\item \textbf{Correctness Guarantee.} Failed trajectories must never receive positive advantage:
\begin{equation}
R(y) < R_{\text{min\_correct}} \implies A(y) \le 0
\end{equation}
where $R_{\text{min\_correct}}$ is a minimum correctness threshold. This condition is meaningful in settings where rewards are not calibrated AND correctness is objective. This constraint then ensures that incorrect behaviors are never reinforced, regardless of group composition.

\item \textbf{Proportional Feedback with Sensitivity to Overestimation.} The baseline should provide informative gradients that are proportional to trajectory quality relative to the true expected reward $\mu^*$, not just the sampled group. Moreover, in light of recent work showing that GRPO exhibits 
exploitative dynamics~\cite{he2025rewardingunlikelyliftinggrpo, 
zhu2025surprisingeffectivenessnegativereinforcement}, the baseline should err toward advantage underestimation rather than overestimation, to prevent premature exploitation.

\item \textbf{Aspirational Drive.} Among correct trajectories ($R(y) \ge R_{\text{min\_correct}}$), 
the baseline should provide competitive pressure to prefer higher-quality 
solutions, enabling continued improvement beyond mere correctness. This creates a two-phase learning dynamic: first, eliminate incorrect behaviors (correctness-seeking); then, refine among correct solutions (quality-seeking).
\end{enumerate}

GRPO's $b_{\text{mean}}$ fails all three: it reinforces failures when 
$b_{\text{mean}} < R(y_f)$, treats over- and underestimation 
symmetrically despite their asymmetric consequences, and provides 
relative comparison among all trajectories without first establishing correctness. We require a baseline that addresses all 
three issues simultaneously.

\subsection{\corpo: Correctness-Relative Policy Optimization}
\label{sec:corpo}

We now present Correctness-Relative Policy Optimization (CoRPO), a simple 
modification to GRPO's baseline that satisfies all three properties, by simply clipping the 
group-mean baseline at a minimum correctness threshold.

A natural way to enforce correctness, when the rewards are ordinal, is to compare rewards against a fixed quality threshold rather than a group-relative baseline. Let $R_{\text{min\_correct}}$ denote the minimum reward for a trajectory to be considered acceptably correct. Any trajectory with reward below this threshold should never receive positive advantage. A purely static baseline, however, lacks competitive pressure among correct solutions and fails to encourage the policy to move from “good” to “optimal.” To reconcile these objectives, we introduce an adaptive baseline that interpolates between a correctness threshold and the GRPO mean baseline.

Given a group of $G$ rollouts $\{y_1, \dots, y_G\}$ sampled from policy $\pi_\theta$, we first compute the standard GRPO baseline using Equation~\ref{eq:grpo_baseline}. We then define the $\corpo$ baseline by clamping this group mean at the correctness threshold:
\begin{equation}
b_{\corpo} = \max\left(R_{\text{min\_correct}},\, b_{\text{mean}}\right).
\end{equation}

The resulting advantage is:
\begin{equation}
A_{\corpo}(y_i) = R(y_i) - b_{\corpo}.
\end{equation}

This formulation yields an adaptive learning signal that operates in two regimes. 

\textbf{A. Correctness-Seeking Regime.}
When the policy performs poorly and a group's average reward falls below the correctness threshold, i.e. $b_{\text{mean}} < R_{\text{min\_correct}}$, the baseline is fixed at $R_{\text{min\_correct}}$. In this regime, $\corpo$ behaves like a static, quality-based baseline. Any failed trajectory $y_f$ with $R(y_f) < R_{\text{min\_correct}}$ is guaranteed to receive negative advantage:
\begin{equation}
A_{\corpo}(y_f) < 0.
\end{equation}
This eliminates the reinforcement of sub-optimal behavior while providing proportional negative feedback based on the severity of failure. By fixing the baseline above $b_{\text{mean}}$, this regime also mitigates the advantage overestimation identified in Section~\ref{sec:challenge1} for samples in group where $b_{\text{mean}} < \mu^*$. 

\textbf{B. Quality-Seeking Regime.}
Once the policy reliably produces correct solutions, i.e $b_{\text{mean}} \ge R_{\text{min\_correct}}$, the baseline transitions to the group mean, $b_{\text{CoRPO}} = b_{\text{mean}}$, recovering GRPO’s relative preference mechanism. Correct trajectories now compete against one another, encouraging the policy to favor higher-quality solutions over merely acceptable ones.

CoRPO's adaptive baseline simultaneously addresses all requirements from 
Section~\ref{sec:ideal_baseline}. 

    (i) $max(.)$ operation ensures 
    $b_{\text{CoRPO}} \ge R_{\text{min\_correct}}$ always, making 
    $A_{\text{CoRPO}}(y_f) \le 0$ for any $R(y_f) < R_{\text{min\_correct}}$.
    
    (ii) When $b_{\text{mean}} < \mu^*$, clipping raises the baseline toward $R_{\text{min\_correct}}$ when $b_{\text{mean}} < R_{\text{min\_correct}}$, 
    reducing advantage overestimation. The asymmetric clipping only 
    from below, accepts modest underestimation to prevent exploitation.
    
    (iii) Once $b_{\text{mean}} \ge R_{\text{min\_correct}}$, 
    the baseline adapts to policy improvements by switching to $b_{\text{mean}}$, maintaining competitive pressure 
    among correct solutions.

Crucially, CoRPO achieves these properties through just a max operation on group-mean baseline, preserving GRPO's computational 
efficiency while eliminating its correctness violations and exploitation tendencies.

\subsection{Addressing Advantage Overestimation with CoRPO}

GRPO yields biased advantage estimates at the level of individual samples due to sampling variance. Although the estimator is unbiased in expectation over all possible groups, it exhibits systematic per-sample bias: some trajectories receive overestimated advantages, while others receive underestimated ones. We argue that advantage overestimation drives exploitative training dynamics, whereas underestimation is comparatively less harmful and may even encourage beneficial exploration. 

GRPO overestimates advantage for a sample whenever $b_{\text{mean}} < \mu^*$ , and vice versa. CoRPO behaves differently depending on where $R_{\text{min\_correct}}$ lies relative to $b_{\text{mean}}$ and $\mu^*$:

\begin{enumerate}
\item \textbf{Case 1: $b_{\text{mean}} < R_{\text{min\_correct}} < \mu^*$} : 
CoRPO reduces overestimation by raising the baseline to $R_{\text{min\_correct}}$.
\item \textbf{Case 2: $b_{\text{mean}} < \mu^* < R_{\text{min\_correct}}$} : 
CoRPO underestimates advantage while GRPO would overestimate.
\item \textbf{Case 3: $\mu^* < b_{\text{mean}} < R_{\text{min\_correct}}$} :
CoRPO increases the magnitude of underestimation relative to GRPO.
\item \textbf{Case 4: $b_{\text{mean}} \geq R_{\text{min\_correct}}$} :
CoRPO behaves identical to GRPO, as the $\max(\cdot)$ operation leads to using $b_{\text{mean}}$ as the baseline.
\end{enumerate}

\subsection{Choosing the Correctness Threshold}

The choice of the correctness threshold $R_{\text{min\_correct}}$ is inherently open, as it arises precisely in settings where the reward function is not well calibrated. CoRPO uses $R_{\text{min\_correct}}$ to inject prior structure into advantage estimation, compensating for this lack of calibration.

A choice of $R_{\text{min\_correct}}$ in such cases can be informed by two considerations. First, it should reflect how a well-calibrated reward would behave. Trajectories that are incorrect under the ground-truth signal should not receive positive advantage (eg, a correct candidate solution receiving a verifier score below the nominal correctness cutoff on a 0–10 scale). Setting $R_{\text{min\_correct}}$ near this calibrated cutoff encourages the policy toward consistent reward interpretation. 

Second, the threshold should account for the empirical reward distribution, balancing the reduction of advantage overestimation against the risk of excessive underestimation. In the regime where $\mu^* \;\le\; R_{\text{min\_correct}} \;\le\; \mu^* + O(\sigma_b)$ , baseline clipping can substantially reduce overestimation while avoiding extreme underestimation. Although $\mu^*$ varies across task difficulty, with $\mu^*_{\text{easy}} > \mu^*_{\text{medium}} > \mu^*_{\text{hard}}$, informing the choice of $R_{\text{min\_correct}}$ using $\mu^*_{\text{medium}}$ allows to reduce advantage overestimation on medium and hard task, while behaving similar to GRPO on most easy tasks. 

Overall, setting $R_{\text{min\_correct}}$ close to the expected correctness threshold under calibrated rewards and $\mu^*_{\text{medium}}$ allows training to operate in the regimes where CoRPO most effectively suppresses overestimation without unduly slowing learning.

%% file: 5.results-v2.tex
\begin{table*}[!ht]
\centering
\small
\setlength{\tabcolsep}{4.5pt}
\begin{tabular}{l l | cc cc | cc cc}
\toprule
& Validation Set
& \multicolumn{4}{c|}{\textbf{In-domain}} 
& \multicolumn{4}{c}{\textbf{Out-of-domain}} \\

\cmidrule(lr){3-10}
& 
& \multicolumn{2}{c}{pass@16} & \multicolumn{2}{c|}{mean@16}
& \multicolumn{2}{c}{pass@16} & \multicolumn{2}{c}{mean@16} \\

Train Set & 
& GRPO & CoRPO & GRPO & CoRPO
& GRPO & CoRPO & GRPO & CoRPO \\
\midrule

\multirow{4}{*}{\textbf{Coding}}
& Overall 
& 87.8 & 87.2 & \textbf{67.1} & 65.5 
& 88.8 & \textbf{90.1} & 60.2 & \textbf{61.7} \\

\cmidrule(lr){2-10}
& \quad Easy 
& 100.0 & 100.0 & \textbf{95.8} & 94.6 
& 100.0 & 100.0 & 94.5 & \textbf{97.3} \\

& \quad Medium 
& 100.0 & 100.0 & \textbf{93.1} & 88.8 
& 100.0 & 100.0 & 78.2 & \textbf{82.3} \\

& \quad Hard 
& 81.7 & 80.9 & 53.4 & 52.4 
& 78.8 & \textbf{81.2} & 35.7 & \textbf{35.9} \\

\midrule
\multirow{4}{*}{\textbf{Math}}
& Overall 
& 89.3 & 88.4 & \textbf{63.0} & 61.5 
& 83.4 & 83.9 & 47.0 & 47.2 \\

\cmidrule(lr){2-10}
& \quad Easy 
& 100.0 & 100.0 & \textbf{95.3} & 93.8 
& 100.0 & 100.0 & \textbf{90.1} & 88.9 \\

& \quad Medium 
& 91.9 & \textbf{97.3} & 75.2 & \textbf{79.1}
& 96.6 & \textbf{100.0} & 62.5 & \textbf{67.2} \\

& \quad Hard 
& \textbf{82.5} & 78.8 & \textbf{41.1} & 37.8 
& 75.6 & 75.6 & 31.3 & 30.9 \\

\bottomrule
\end{tabular}
\caption{Performance (\%) on in-domain and out-of-domain evaluations, stratified by 
difficulty. Bold indicates improvements $>1\%$}
\label{tab:grpo-corpo}
\vspace{-18pt}
\end{table*}

\section{Experiment Results}

We empirically validate our claims by training an LLM-verifier for math and coding tasks with RLVR. We compare models trained with GRPO and CoRPO on in-domain (ID) and out-of-domain (OOD) validation sets, showing that CoRPO consistently achieves superior generalization. We further analyze the underlying training dynamics, demonstrating systematic advantage overestimation in GRPO across task difficulties, and show that CoRPO enforces more conservative updates via consistent negative reinforcement of incorrect behaviors. This mitigates distribution sharpening, and induces implicit curriculum learning, leading to improved cross-domain generalization.

\subsection{Training Setup}
\label{sec:trainsetup}

We train explanatory verifiers for coding and math using RLVR. All models are initialized from Qwen3-8B~\cite{yang2025qwen3} and trained using either GRPO or CoRPO, following the VeRL implementation~\cite{verl}. We train separate models for coding and math tasks.

\textbf{Task Description.} Each problem instance consists of a question $Q$ and two candidate responses $(C_A, C_B)$. The policy outputs ratings $V = (v_A, v_B) \in \{-2, -1, 1, 2\}$, representing its confidence in the correctness of each response. Although ratings are assigned to each response individually, the pairwise formulation enables the verifier to leverage relative evidence between competing solutions, promoting more discriminative and reliable verification. Policy output is guided by the following instruction in user prompt:

\begin{tcolorbox}[colback=orange!10, colframe=orange!70, 
                  rounded corners, boxrule=0.3mm, arc=4mm]
Given a question and multiple responses from the Assistant, you need to identify how likely it is that the given responses are correct. Each score can be one of $\{-2, -1, 1, 2\}$, with higher values indicating greater confidence. For example, a score of -2 means you are very confident that the response is incorrect, \
a score of -1 means likely incorrect, 1 means likely correct, and 2 means you are very confident the response is correct.
\end{tcolorbox}

\paragraph{Reward Computation.}

The reward is defined based on the difference in predicted ratings, $(v_B - v_A)$, and a target label of $\mathbb{I}(y_B > y_A)$, where $y_i \in \{0,1\}$ is the ground truth label of corresponding response. Formally:
\[
\text{diff\_rating} =
\begin{cases}
v_B - v_A, & \text{if C}_B \text{ is correct} \\
v_A - v_B, & \text{if C}_A \text{ is correct}
\end{cases}
\]

We further map the signed rating difference to predefined values, yielding a coarse ordinal reward scheme:
\[
\text{Reward} =
\begin{cases}
1.0,  & \text{if } \text{diff\_rating} \in [3,4] \\
0.5,  & \text{if } \text{diff\_rating} \in [1,2] \\
-1.0, & \text{if } \text{diff\_rating} \in [-2,0] \\
-2.0, & \text{if } \text{diff\_rating} \in [-4,-3] \\
-2.0, & \text{otherwise}
\end{cases}
\]

We train on two datasets: coding and math. More details on task description are in Appendix~\ref{sec:task} and data processing in Appendix~\ref{sec:data}. The formal algorithm is described in Appendix~\ref{sec:algo}. We also experiment with different ordinal reward granularity, details of which are included in Appendix~\ref{sec:ordinal}. 

\textbf{Choosing $R_{min\_correct}$}.
For the verification task, correctness is defined by whether the policy predicts the correct direction of the rating difference. Specifically, when response $C_B$ is correct, we require $v_B > v_A$. Under this criterion, we set the minimum correctness threshold to $R_{min\_correct} = 0$. As shown in the following analysis, this choice also prevents advantage overestimation on easy and medium tasks.

\subsection{Main Result: CoRPO Enables Robust Cross-Domain Transfer}
\label{subsec:downstream-accuracy}

We compare downstream accuracy of models trained with GRPO and \corpo\ . Table~\ref{tab:grpo-corpo} reports \texttt{pass@16} and \texttt{mean@16} on ID and OOD validation tasks, for models trained on coding and math datasets. All evaluations are done at 16k MSL with temp=0.3.  

CoRPO consistently outperforms GRPO on OOD tasks across both 
training regimes. When trained on coding, CoRPO achieves 
90.1\% vs. GRPO's 88.8\% on math tasks - despite math being easier. This reverse-difficulty transfer demonstrates that CoRPO 
learns generalizable reasoning patterns rather than domain-specific heuristics. 
GRPO's in-domain advantage appears primarily in mean@16 (67.1\% vs. 65.5\%) 
rather than pass@16 (87.8\% vs. 87.2\%), confirming distribution sharpening rather than diverse exploration. 

\begin{figure}[h!]
    \centering
    \includegraphics[width=\linewidth]{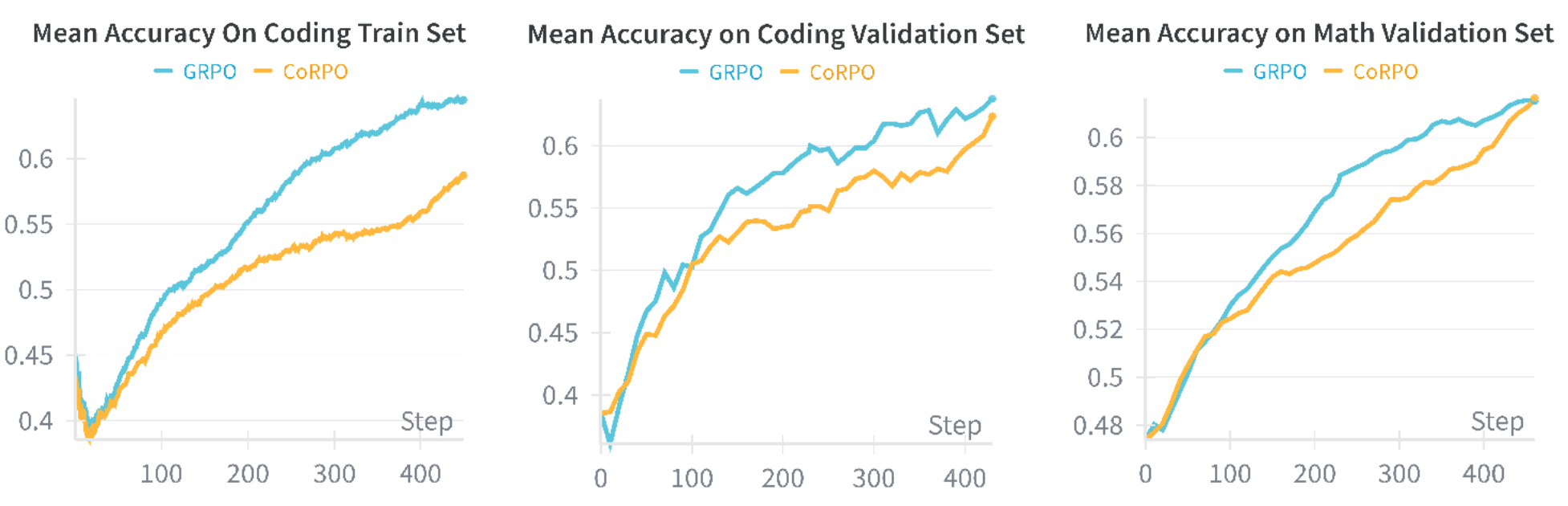} 
    \caption{CoRPO matches or outperforms GRPO on validation sets, even with a lower training accuracy on training set.}
    \label{fig:id-ood-validation}
    \vspace{-10pt}
\end{figure}

Figure~\ref{fig:id-ood-validation} compares training and validation accuracy as the training progressed. GRPO achieves higher training accuracy than \corpo\ throughout training. Validation accuracy on ID tasks is comparable for both methods. In contrast, \corpo\ exhibits superior out-of-domain transfer: validation accuracy on OOD steadily improves and surpasses GRPO toward the end of training.

\begin{figure*}[h!]
    \centering
     \includegraphics[width=\linewidth]{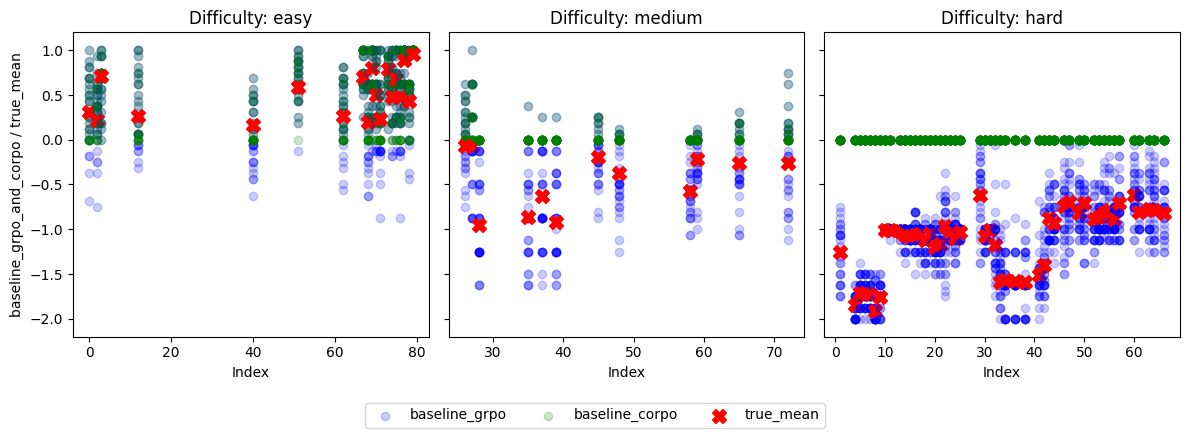} 
    \caption{
Baseline behavior of GRPO and CoRPO across task difficulty.
\textbf{Easy:} CoRPO largely coincides with GRPO as the sample mean reward typically exceeds the correctness threshold. In cases where the sample mean severely underestimates the true mean and falls below the correctness threshold, CoRPO clips the baseline upward. 
\textbf{Medium:} The most critical regime, where the true mean lies near the correctness threshold; CoRPO updates the baseline for many samples, substantially reducing advantage overestimation relative to GRPO. 
\textbf{Hard:} All samples have true mean reward below the correctness threshold. While this induces conservative (underestimated) advantages, it guarantees the absence of advantage overestimation, preventing reinforcement of incorrect trajectories.
}
    \label{fig:baseline-correction}
    \vspace{-15pt}
\end{figure*}

To understand the source of this performance gap, we stratify results by problem 
difficulty. GRPO's ID gains concentrates heavily on medium-difficulty problems 
(93.1\% vs. 88.8\% mean@16 when trained on coding), but this advantage 
disappears, and reverses, on OOD evaluation (78.2\% vs. 82.3\%). This suggests that GRPO’s optimization dynamics favor improvements on a narrow subset of ID examples, which do not translate to robust performance under distribution shift.

\subsection{Conservative Advantage Assignment with \corpo}
\label{sec:grpoflaw}
CoRPO is able to achieve better cross-domain generalization by being effective in two complementary regimes. 

\textbf{Advantage overestimation.}
First, CoRPO mitigates advantage overestimation by shifting the baseline upward, thereby reducing overconfident updates and promoting more stable training dynamics. To examine this effect across task difficulty levels, we sample 128 rollouts for a subset of 80 tasks from the training dataset, using the mean of these 128 rollouts as the true expected baseline. We then bootstrap 30 samples of size 8, reflecting the group sizes used during RL training, to compute GRPO and CoRPO baselines. Figure~\ref{fig:baseline-correction} shows the distribution of estimated baselines relative to the true expected baseline for each task index across difficulty levels. GRPO overestimates advantage for nearly half of the samples across difficulties, as its estimated baseline tends to lie below the true mean. By contrast, CoRPO’s upward-shifted baseline substantially reduces this overestimation. For medium- and hard-difficulty tasks, CoRPO entirely eliminates advantage overestimation. On easy tasks, the standard error of advantage estimation,
$\sqrt{\sum((R - \mu^*) - (R - \hat{b}))^2}$,
is reduced by 30\% relative to GRPO. This correction is a key mechanism by which CoRPO avoids overfitting to training-domain patterns and improves generalization.

\textbf{Failed trajectories receiving positive advantage.}
Second, for rollout groups in which a large fraction of samples fall below the correctness threshold, CoRPO corrects GRPO’s failure mode by (i) preventing rollouts with reward below the correctness threshold from receiving positive advantage, and (ii) limiting the magnitude of positive advantage assigned to the few rollouts whose rewards exceed the correctness threshold. Together, these effects suppress extreme positive advantages and prevent overly strong gradient updates driven by sparse correct samples.

\begin{figure}[h!]
    \centering
    \includegraphics[width=\linewidth]{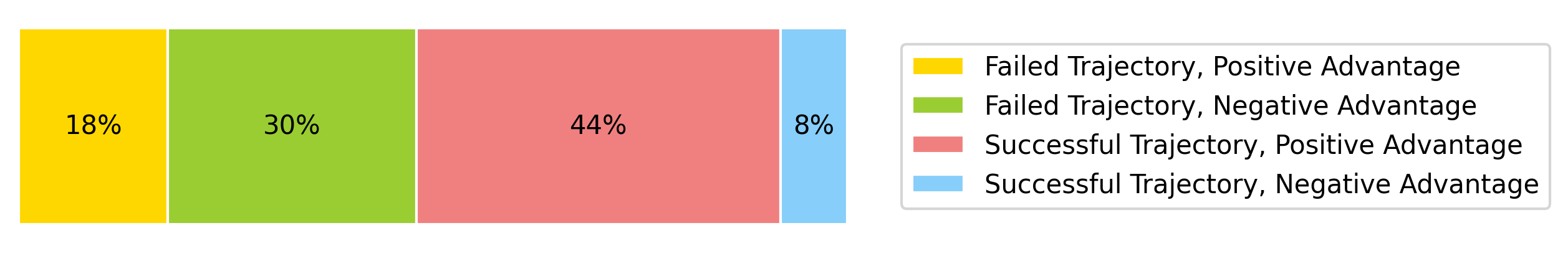}
    \caption{Advantage sign versus trajectory status for a representative GRPO batch. The 18\% mass assigned to failed trajectories with positive advantage occurs when $b_{mean} < R(y_f) < 0$ .}
    \label{fig:grpo-flaw-pie}
    \vspace{-15pt}
\end{figure}

To validate our hypothesis from Section~\ref{sec:challenge2} that GRPO baseline, $b_{\text{mean}}$, assigns positive advantage to failed trajectories. We analyze a representative batch of rollouts from an early stage of GRPO training. Figure~\ref{fig:grpo-flaw-pie} shows that a non-trivial fraction of rollouts fall into the \textbf{Failed Trajectory, Positive Advantage} category, providing direct empirical evidence that GRPO can reinforce suboptimal behavior and violate the correctness guarantee.

\subsection{Training Dynamics of \corpo\ vs GRPO}
\label{sec:traindyn}

\begin{table*}[h!]
\centering
\small
\setlength{\tabcolsep}{8pt}
\begin{tabular}{l l | cc | cc}
\hline
Domain & Task 
& \multicolumn{2}{c|}{Mid-training} 
& \multicolumn{2}{c}{End of training} \\
& 
& GRPO & CoRPO 
& GRPO & CoRPO \\
\hline
\multirow{3}{*}{In-domain (Coding)} 
& Easy 
& 100.0 & 100.0 
& 100.0 & 100.0 \\
& Medium 
& 100.0 & 96.6 
& 100.0 & 100.0 \\
& Hard 
& 87.0 & 75.6 
& 81.7 & 80.9 \\
\hline
\multirow{3}{*}{Out-of-domain (Math)} 
& Easy 
& 100.0 & 100.0 
& 100.0 & 100.0 \\
& Medium 
& 100.0 & 100.0 
& 100.0 & 100.0 \\
& Hard 
& 66.2 & 66.2 
& 78.8 & 81.2 \\
\hline
\end{tabular}
\caption{
Pass@16 of GRPO and CoRPO at mid-training and end of training. CoRPO exhibits slower early progress on hard in-domain tasks but matches GRPO by convergence and achieves stronger out-of-domain performance, particularly on hard problems.
}
\label{tab:grpo_corpo_curriculum}
\vspace{-20pt}
\end{table*}

\subsubsection{Learning from Negative Reinforcement}

\begin{figure}[h!]
    \centering
    \includegraphics[width=\linewidth]{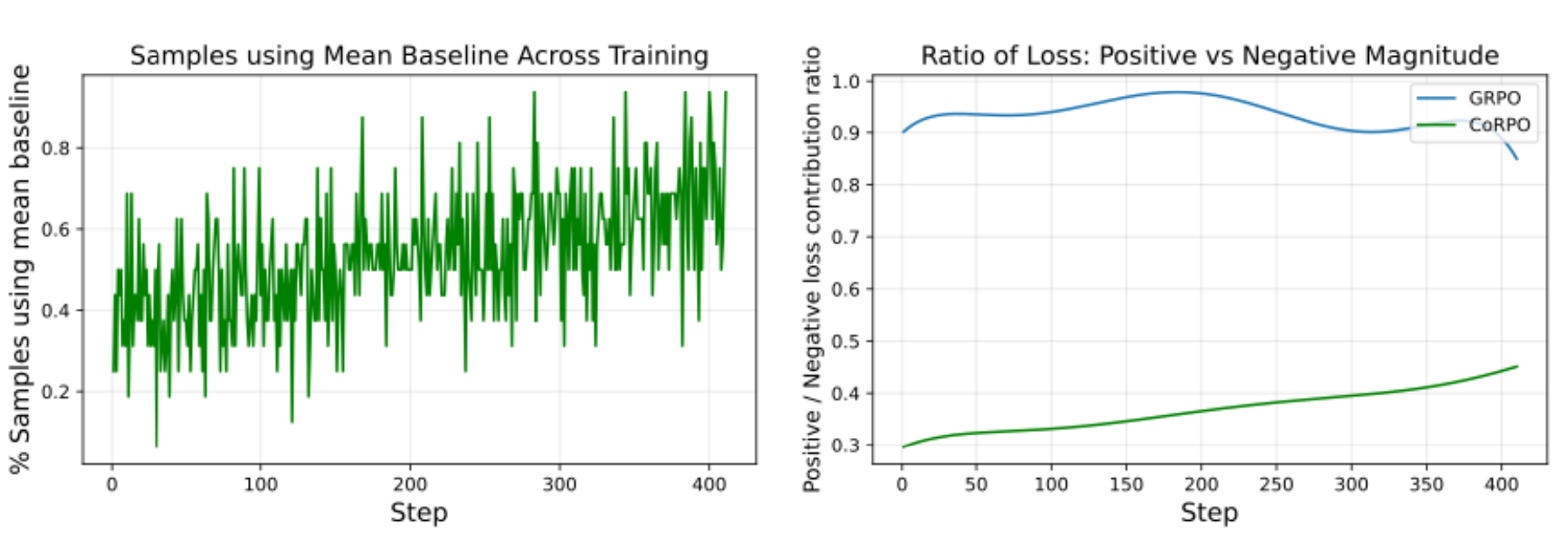} 
    \caption{(Left) Proportion of samples using $b_{mean}$ for advantage calculation goes up from 30\% to 80\% as the training progresses with \corpo. (Right) Ratio of negative to positive advantage, $r_{loss}$, steadily increases for \corpo\ but stays high at 1 throughout training for GRPO. }
    \label{fig:lossratio}
    \vspace{-10pt}
\end{figure}

Figure~\ref{fig:lossratio} (left) shows that early in training, only $\sim$30\% of $\corpo$\ samples use $b_{mean}$; the rest are clipped to $R_{\text{min\_correct}}$, ensuring failed trajectories receive negative advantage. Figure~\ref{fig:lossratio} (right) reports the ratio of loss contributions from positive and negative advantages, $r_{\text{loss}}=\frac{\sum \mathrm{Loss}_{A(y)>0}}{\sum \mathrm{Loss}_{A(y)<0}}$. In CoRPO, $r_{\text{loss}}\approx0.3$ at the beginning of training indicates negative reinforcement dominates, and $r_{\text{loss}}$ continues to improve during training, reflecting a transition toward reinforcing higher-quality solutions. In contrast, GRPO maintains $r_{\text{loss}}\approx1$ throughout, learning equally from positive and negative advantages, which can cause premature exploitation early in training. 

Prior work shows GRPO suffers from \emph{rank bias}\cite{he2025rewardingunlikelyliftinggrpo}, concentrating probability mass on already-likely solutions and reducing exploration. CoRPO mitigates this by learning primarily from negative reinforcement early, uniformly reinforcing correct trajectories regardless of initial likelihood (see Appendix \ref{sec:rankbias}).


\subsubsection{\corpo\ induces implicit curriculum learning} 

We demonstrate CoRPO's and GRPO's performances during the training process in Table \ref{tab:grpo_corpo_curriculum}. For ID tasks, CoRPO progresses slower on hard tasks early but matches GRPO on easy and medium tasks, and reaches parity on hard tasks at convergence. In contrast, GRPO’s performance on hard tasks degrades from 87\% at mid-training to 81.7\% by the end of training. For OOD tasks, CoRPO already matches GRPO at mid-training and consistently outperforms it by the end of training across all difficulty levels. Notably, the CoRPO's gains are most pronounced on hard problems.

Overall, these results suggest that CoRPO implements an implicit curriculum learning. By blocking suboptimal trajectories from receiving positive advantage early, it suppresses noisy updates and gradually introduces harder examples once reliably solved. The result is a temporary lag on hard ID tasks but consistently stronger OOD performance. In effect, CoRPO prioritizes correctness and stability first, difficulty later, yielding more robust and generalizable learning.

\begin{figure}[h!]
    \centering
    \includegraphics[width=\linewidth]{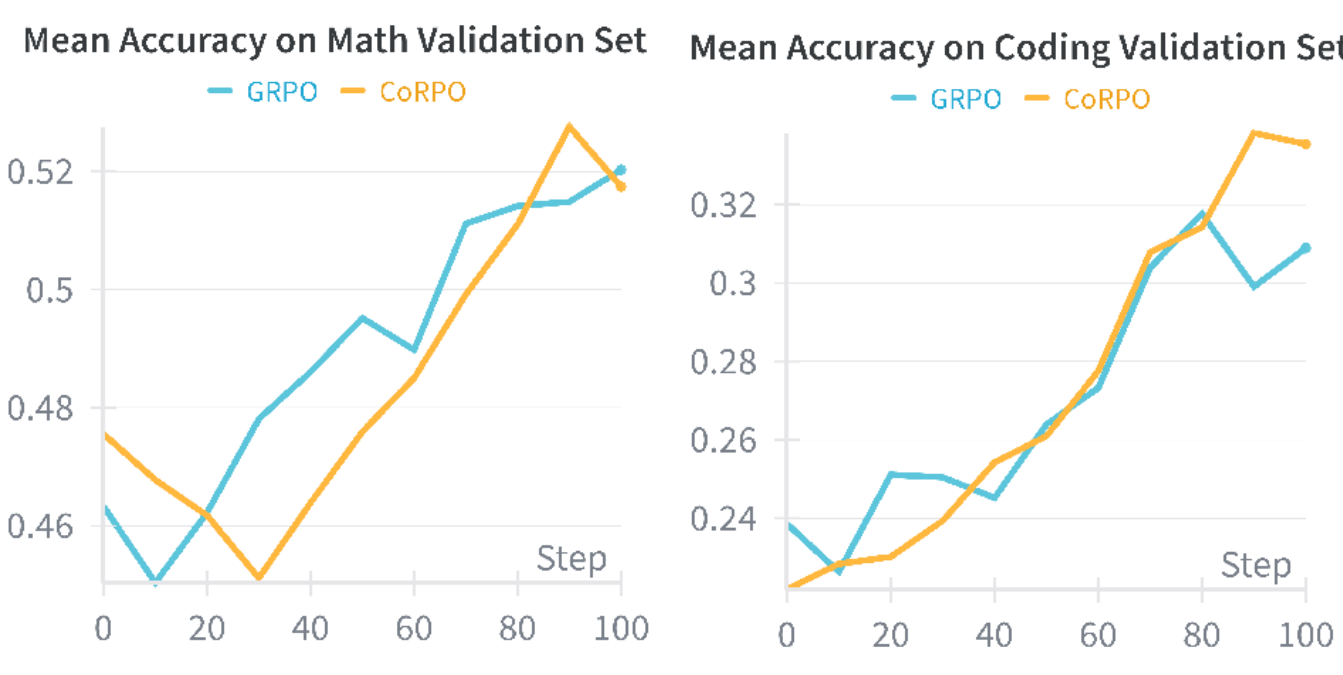} 
    \caption{When training with small group size of 4, CoRPO surpasses GRPO OOD performance by as early as 100 steps.}
    \label{fig:math-small-group}
    \vspace{-10pt}
\end{figure}

\subsection{Robustness to Group Size}
GRPO's mean baseline becomes increasingly biased with fewer rollouts, as the sample mean poorly estimates the true expected reward. We test CoRPO's robustness by training with only n=4 rollouts per prompt, half the standard group size. Figure \ref{fig:math-small-group} shows that CoRPO outperforms GRPO after just 100 training steps with n=4, demonstrating that a clipped baseline provides a more stable learning signal even under high-bias conditions. This suggests CoRPO may be particularly valuable in resource-constrained settings where fewer rollouts per prompt are necessary.

%% file: 6.conclusion.tex
\section{Conclusion}

GRPO’s group-mean baseline can overestimate advantages and positively reinforce incorrect trajectories. This leads to reinforcing suboptimal behaviors and can cause premature exploitation. CoRPO addresses this by enforcing a minimum correctness threshold, ensuring incorrect trajectories never receive positive advantage while retaining GRPO’s efficiency. By optimizing relative to correctness, CoRPO encourages robust learning without amplifying relatively-better but incorrect samples. This baseline clipping also mitigates advantage overestimation and promotes effective exploration.

Empirically, CoRPO improves cross-domain generalization. Models trained on coding tasks outperform GRPO on downstream math tasks, indicating it fosters robust, transferable reasoning. Additional analysis reveals that CoRPO effectively mitigates distribution sharpening and rank bias, and induces implicit curriculum learning.


\newpage

%% file: 7.appendix.tex
\newpage
\appendix
\section{Task Description}
\label{sec:task}

We train a pairwise LLM verifier using reinforcement learning. Each input consists of a question and two responses sampled from Qwen3-8B. Although the model assigns ratings to each response individually, the task is structured to encourage comparative reasoning between the two responses. The desired outcome is a higher rating for the correct response, which is reflected in the reward design based on the difference between the two predicted ratings.

\begin{tcolorbox}[
  colback=white,
  colframe=black,
  boxrule=0.8pt,
  arc=1mm,
  breakable
]
\textbf{Prompt}

\small
Given a question and multiple responses from the Assistant, you need to
identify how likely it is that the given responses are correct.
Each score can be one of \{-2, -1, 1, 2\}, with higher values indicating
greater confidence. For example, a score of -2 means you are very confident that the response is incorrect, \
a score of -1 means likely incorrect, 1 means likely correct, and 2 means you are very confident the response is correct.

Before scoring, please analyze step by step. Your scoring should be as
strict as possible.

\texttt{\#\#\#\# Question Begin \#\#\#\#} \\
\texttt{\{question\}} \\

\texttt{\#\#\#\# Responses to be Scored \#\#\#\#} \\
\texttt{[Begin Response 1]} \\
\texttt{\{response1\}} \\
\texttt{[The End of Response 1]} \\

\texttt{[Begin Response 2]} \\
\texttt{\{response2\}} \\
\texttt{[The End of Response 2]} \\

\texttt{\#\#\#\# Output Format Requirements \#\#\#\#} \\
Analysis: \textless step-by-step comparison\textgreater \\
Scores: \textbackslash boxed\{x, x\}
\end{tcolorbox}


\begin{tcolorbox}[
  colback=white,
  colframe=black,
  boxrule=0.8pt,
  arc=1mm
]
\textbf{Model Output}

\small
\texttt{
<think> reasoning </think>
....\\
\#\#\# **Final Answer**
\textbackslash boxed\{-1, 2\}}
\end{tcolorbox}

Let $\hat{s} = (\hat{s}_1, \hat{s}_2)$ denote the predicted confidence scores, and
$s = (s_1, s_2) \in \{0,1\}$ denote the ground-truth verification labels.
We define $s_i = 1$ if $\text{response}_i$ is correct, and $s_i = 0$ otherwise. A few example reward calculations based on model output are shared below. \\
\textbf{Example 1:}
\[
s = \{0, 1\}, \quad \hat{s} = \{-1, 2\}
\]
\[
\text{diff\_rating} = 2 - (-1) = 3
\]
\[
\text{Reward} = 1 
\]

\textbf{Example 2:}
\[
s = \{1, 0\}, \quad \hat{s} = \{-1, 2\}
\]
\[
\text{diff\_rating} = -1 - 2 = -3
\]
\[
\text{Reward} = -2 
\]

\section{Failed trajectory getting a positive advantage}
\label{sec:failedtrajectory}
\begin{tcolorbox}[title=Example 1: Positive Advantage to Failed Trajectory with GRPO]

\textbf{Question:} \\
\emph{Scoring Instructions omitted for brevity} \\
\#\#\#\# Question Begin \#\#\#\# \\
\texttt{\{python coding question\}} \\
\#\#\#\# Responses to be Scored \#\#\#\# \\
\texttt{[Begin Response 1]} \\
\texttt{\{response1\}} \\
\texttt{[The End of Response 1]} \\

\texttt{[Begin Response 2]} \\
\texttt{\{response2\}} \\
\texttt{[The End of Response 2]} \\

\#\#\#\# Output Format Requirements \#\#\#\# \\
Analysis: \textless step-by-step comparison\textgreater \\
Scores: \textbackslash boxed\{x, x\} \\[6pt]

\textbf{Ground Truth Verifications:} \\
Response 1 is correct, Response 2 is incorrect \\[6pt]

\textbf{Model output:} \\  
\emph{(truncated for brevity)}.... \\
Response 1 : The binary search logic is flawed, and the approach may not work for all cases. \\
Response 2 : The logic is correct but may not pass due to inefficiency in large scenarios.  \\
\#\#\# Final Answer
\boxed{-1, 1}
\\[6pt]

\textbf{Reward of the sample:} \\
diff rating = -1 - (1) = -2 \\
reward = -1 \# As per computations in Appendix A  \\[6pt]

\textbf{Group Mean}: -1.125 \\
\textbf{Advantage GRPO}: -1 - (-1.125) = +0.125 \\
\textbf{Advantage CoRPO}: -1 - 0 = -1

\end{tcolorbox}

\section{Data Processing And Training Hyperparameters}
\label{sec:data}
The coding dataset is curated from Codeforces~\cite{li2023taco} and LeetCode~\cite{xia2025leetcodedatasettemporaldatasetrobust}, while the math dataset is drawn from NuminaMath~\cite{li2024numinamath} after Skywork-OR1~\cite{he2025skywork} filtering. For each question, we sample multiple reasoning trajectories generated by Qwen3-8B to obtain both correct and incorrect solutions. For math questions, we obtain ground truth verifications using math-verify. For coding questions, we run the generated code against the set of provided test cases to obtain the ground label. A code is considered correct if it passes all the given test cases, and incorrect otherwise. 

To formulate the input prompts for training verifier, we first remove content within \texttt{<think>} tags from the responses and discard samples exceeding 4,096 input tokens. To stabilize the absolute scale of ratings, we additionally include a small subset of single-response judgment examples $(Q, R_A \rightarrow v_A)$. The final training sets contain 4,890 coding samples and 3,507 math samples. We evaluate on in-domain and out-of-domain validation sets constructed from $(Q, R_A, R_B)$ tuples, with balanced correctness orderings. The coding validation set contains 196 samples and the math validation set contains 157 samples.

 For coding, we use a maximum sequence length of 16,384 with 8 rollouts per prompt and a global batch size of 512. For math, we use a maximum sequence length of 8,192 with 16 rollouts per prompt and a global batch size of 256. We apply dynamic filtering~\cite{yu2025dapoopensourcellmreinforcement} to discard rollouts with zero reward variance and perform a single gradient update per batch to ensure strictly on-policy training~\cite{chen2025acereason}. We set both the KL and entropy coefficients to 0.0. The learning rate is $1\times10^{-6}$ with a 10-step linear warmup, and rollouts are generated with temperature and top-$p$ set to 1.0.

\input{algorithm1}

\section{Algorithm}
\label{sec:algo}

The algorithm used for training is described in Algorithm~\ref{alg:verifier_corporl}.

\section{Impact of Reward Granularity}
\label{sec:ordinal}
We compare two ordinal rating schemes: a 4-point scale with outputs in ${-2,-1,1,2}$, and a 10-point scale with integer outputs in ${0,1,\dots,10}$. This comparison allows us to study the effect of reward granularity on training stability and generalization performance. We observed that using a coarser 4-point scale yields a dramatic improvement in the stability and magnitude of the learning signal. As shown in Table~\ref{tab:corpotrainhp}, the mean@16 score jumps significantly (e.g., rising from $0.37$ to $0.56$ for In-Domain tasks). This suggests the policy model is receiving higher quality signals, avoiding the optimization pitfalls of vanishing gradients associated with minor partial credits. While this coarser signal caused a slight regression in In-Domain pass@16 (dropping from $0.85$ to $0.83$), it continues to maintain/improve performance on out-of-domain tasks. Motivated by these results, we continued training with 4-point scale to compare GRPO and \corpo\ for all future experiments.

\begin{table}[t!]
\centering
\small
\begin{tabular}{lcccc}
\toprule
 & \multicolumn{2}{c}{\textbf{10-point}} 
 & \multicolumn{2}{c}{\textbf{4-point}} \\
\cmidrule(lr){2-3} \cmidrule(lr){4-5}
 & pass@16 & mean@16 & pass@16 & mean@16 \\
\midrule
In-Domain  & 85.0 & 37.0 & 83.0 & 56.0 \\
OOD   & 82.0 & 50.0 & 82.0 & 53.0 \\
\bottomrule
\end{tabular}
\caption{Performance (\%) of CoRPO under different scoring schemes on in-domain and out-of-domain tasks.}
\label{tab:corpotrainhp}
\end{table}

\section{Quantifying the Rank Bias in GRPO vs \corpo}
\label{sec:rankbias}
\begin{figure}[h]
    \centering
    \includegraphics[width=0.9\linewidth]{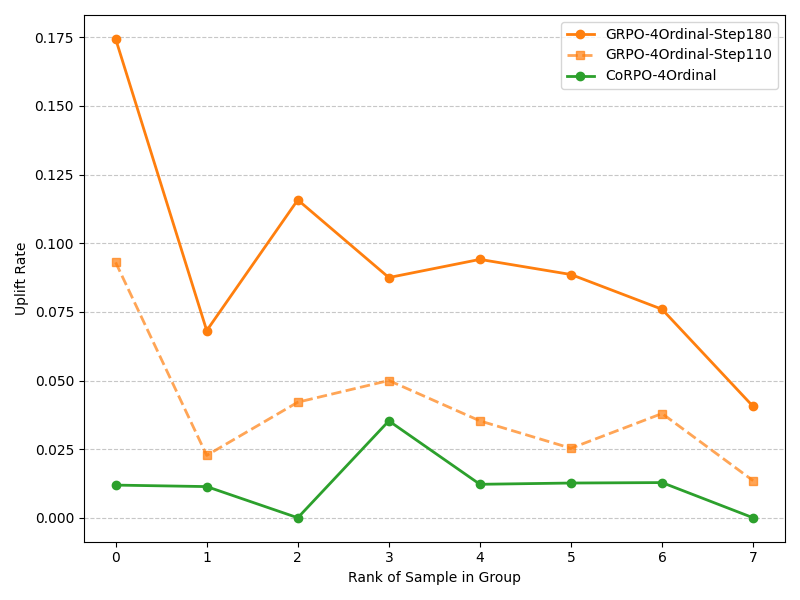} 
    \caption{\textbf{Impact of GRPO vs. \corpo\ on policy distribution dynamics.} The X-axis ranks correct solutions by their initial likelihood (High → Low). The Y-axis shows the Uplift Ratio (post-training probability / pre-training probability of correct solutions). GRPO exhibits distribution sharpening, disproportionately reinforcing high-probability trajectories as training progresses (Step 110 → 180). In contrast, \corpo\ applies uniform reinforcement independent of starting probability. }
    \label{fig:unlikely}
\end{figure}

We quantify the effect of distribution sharpening via \textbf{uplift rate analysis} in Figure~\ref{fig:unlikely}, which measures the change in probability mass assigned to correct solutions from the initial to the final policy. GRPO increasingly favors high-probability solutions as training progresses, while \corpo\ exhibits near-uniform uplift across the likelihood spectrum. This enables extended training without overfitting to dominant training patterns.

We also conducted experiments with updating the default weight decay param from 0.1 to 0 to observe if CoRPO does not need regularization. Our claims on generalizability are further validated by the results showing CoRPO learns even better when using weight decay = 0. More details in Appendix \ref{sec:weightdecay} .

\section{Impact of Removing Weight Decay (WD=0)}
\label{sec:weightdecay}
While training with \corpo\ ,we observe that removing weight decay (WD = 0) leads to faster learning, with in-domain pass@16 and out-of-domain pass@16 and mean@16 reaching a new performance frontier, surpassing GRPO at the same training steps (Table~\ref{tab:corpo4point-wd}). This suggests that CoRPO’s emphasis on negative reinforcement provides sufficient intrinsic regularization, reducing the need for explicit weight decay. By eliminating the dampening effect of weight decay, policy gradients are able to drive the model toward higher-performing solutions more rapidly.

Over longer training however, WD = 0.1 closes much of the gap, particularly in mean@16, although pass@16 on out-of-domain tasks remains higher with WD = 0. This direction needs to be further studied to better understand training dynamics. All experiments reported in the main section therefore use WD = 0.1 for both GRPO and CoRPO to ensure a fair and conservative comparison.

\begin{table}[t!]
\centering
\renewcommand{\arraystretch}{1.2} 
\small
\begin{tabular}{lcccc}
\toprule
 & \multicolumn{2}{c}{\textbf{WD=0.1}} 
 & \multicolumn{2}{c}{\textbf{WD=0}} \\
\cmidrule(lr){2-3} \cmidrule(lr){4-5}
 & pass@16 & mean@16 & pass@16 & mean@16 \\
\midrule
In-Domain  & 83.0 & 56.0 & 89.0 & 59.0 \\
OOD    & 82.0 & 53.0 & 84.0 & 55.0 \\
\bottomrule
\end{tabular}
\caption{Effect of weight decay (WD) on \corpo\ 4-point performance at 200 training steps. Results are reported separately for in-domain and out-of-domain (OOD) tasks when training on coding dataset.}
\label{tab:corpo4point-wd}
\end{table}

%% file: algorithm1.tex
\begin{algorithm}
\caption{RL objective for pairwise LLM verifier}
\label{alg:verifier_corporl}
\begin{algorithmic}
\REQUIRE Dataset $\mathcal{D}=\{x_1, ..., x_N\}$. Each sample $x_i=(p_i, q_i, h_{A,i}, h_{B,i})$, where $p_i$ is prompt, $q_i$ is the question, and $h_{A,i},h_{B,i}$ is a pair of correct/incorrect responses.
\REQUIRE Policy model $\pi_\theta$. Each of the model rollout generation $\pi_\theta(x_i)$ contains two scores $(s_{A,i}, s_{B,i})$ where each score corresponds to a response in the sample. The score $s \in \mathbb{Z}$ in the range $[R_{min}, R_{max}]$. 
\\

\FOR{each sample $x_i=(p_i, h_{A,i}, h_{B,i})$ in $\mathcal{D}$}
    \STATE Get $G$ model rollouts $\pi_\theta(x_i)=[y_{i,1}, ..., y_{i,G}]$
    \FOR{each rollout $y_{ij}=(s_{A,ij}, s_{B,ij})$} 
    
        \STATE $(s^+_{ij}, s^-_{ij})\gets (s_{A,ij},s_{B,ij})$ if $h_{A,i}$ is the correct response else $(s_{B,ij},s_{A,ij})$
        \STATE $\Delta_{ij} \gets s^+_{ij} - s^-_{ij}$
    \ENDFOR
    \\
    
    \STATE Calculate baseline $b_i$ for $\pi_\theta(x_i)$
    \STATE In GRPO, $b_i$ is just the mean $b_{mean,i}=\frac{1}{G}\sum_j\Delta_{ij}$
    \STATE For CoRPO, $b_{corpo}=max(R_{min\_correct}, b_{mean,i})$
    \\

    \FOR {each rollout $y_{ij}$} 
        \STATE $A_{ij} \gets f(\Delta_{ij}) - b_i$
    \ENDFOR
    \\

    \STATE $J_i(\theta)=\mathbb{E}_{y\sim\pi_\theta(x_i)}[A(x_i,y)]$
    
    \STATE $\nabla_\theta J_i(\theta) = \mathbb{E}_{y\sim \pi_\theta(x_i)}\!\left[A(x_i,y)\,\nabla_\theta \log \pi_\theta(y|x_i)\right]$ 
    \\
    
    \STATE  $\theta \gets \theta + \eta \nabla_\theta J(\theta)$ 
    
\ENDFOR
\end{algorithmic}
\end{algorithm}